\documentclass{article}

\usepackage{times}
\usepackage{graphicx} % more modern
\usepackage[english]{babel}

\usepackage[caption=false]{subfig}

\usepackage{natbib}

\usepackage{algorithm}
\usepackage{algorithmic}

\usepackage{amsmath}
\usepackage{amssymb}
\usepackage{amsfonts}
\usepackage{listings}

\newcommand{\R}{\mathbb{R}}
\newcommand{\C}{\mathbb{C}}

\newcommand{\SO}[1]{\ensuremath{\operatorname{SO}(#1)}}

\usepackage{hyperref}

\usepackage[accepted]{icml2015}

\icmltitlerunning{Harmonic Exponential Families}

\begin{document} 

\twocolumn[
%\icmltitle{Bayesian Analysis on Compact Groups and Homogeneous Spaces}
%\icmltitle{Efficient Learning and Inference for Harmonic Exponential Families on Compact Homogeneous Manifolds and Lie Groups}
\icmltitle{Harmonic Exponential Families on Manifolds}
  
% unless you've provided the [accepted] option to the icml2015
\icmlauthor{Taco S. Cohen}{t.s.cohen@uva.nl}
\icmladdress{University of Amsterdam}
\icmlauthor{Max Welling}{m.welling@uva.nl}
\icmladdress{
  University of Amsterdam \\
  University of California Irvine \\
  Canadian Institute for Advanced Research}

\icmlkeywords{keywords here}

\vskip 0.3in
]

\begin{abstract}
  In a range of fields including the geosciences, molecular biology, robotics and computer vision, one encounters problems that involve random variables on manifolds.
  Currently, there is a lack of flexible
  probabilistic models on manifolds that are fast and easy to train.
  We define an extremely flexible class of exponential family distributions on manifolds such as the torus, sphere, and rotation groups,
  and show that for these distributions the gradient of the log-likelihood can be computed efficiently using a non-commutative generalization of the Fast Fourier Transform (FFT).
  We discuss applications to Bayesian camera motion estimation (where harmonic exponential families serve as conjugate priors), and modelling of the spatial distribution of earthquakes on the surface of the earth.
  Our experimental results show that harmonic densities yield a significantly higher likelihood than the best competing method, while being orders of magnitude faster to train.  
\end{abstract}

\section{Introduction}

Many problems in science and engineering involve random variables on manifolds.
In the geosciences, for example, one deals with measurements such as the locations of earthquakes and weather events on the spherical surface of the earth.
In robotics and computer vision, unobserved Lie group transformations such as rotations and rigid-body motions play an important role in motion understanding, localization and alignment problems.
Classical probabilistic models cannot be applied to data on manifolds because these models do not respect the manifold topology (having discontinuities at the value $0 \equiv 2 \pi$ of a circular variable, for instance), and because they are not equivariant (a manifold-preserving transformation of the data would take the distribution outside the model family).
Among manifold distributions there are currently none that are both flexible and efficiently trainable.

In this paper we study a very flexible class of densities on compact Lie groups (such as the group of rotations in two or three dimensions) and homogeneous spaces (such as the circle, torus, and sphere).
We refer to these densities as harmonic exponential families, because they are based on a generalized form of Fourier analysis known as non-commutative harmonic analysis \cite{Chirikjian2001}.
Specifically, the sufficient statistics of these families are given by functions that are analogous to the sinusoids on the circle.

We show that the moment map \cite{Wainwright2007} that takes the natural parameters of an exponential family and produces the moments of the distribution can be computed very efficiently for harmonic exponential families using generalized Fast Fourier Transform (FFT) algorithms.
This leads directly to a very efficient maximum-likelihood estimation procedure
that is applicable to any manifold for which an FFT algorithm has been developed,
and that enjoys the convexity and convergence properties of exponential families.

We apply the harmonic exponential family of the sphere to the problem of modeling the spatial distribution of earthquakes on the surface of the earth.
Our model significantly outperforms the best competing method (a mixture model), while being orders of magnitude faster to train.

Harmonic exponential families also arise naturally as conjugate priors in a Bayesian framework for estimating transformations from correspondence pairs.
In this case the points on the manifold (the transformations) are not observed directly.
Instead we see a pair of vectors $x$ and $y$ --- images, point clouds, or other data --- that provide information about the transformation that produced one from the other.
If we have a prior $p(g)$ over transformations and a likelihood $p(y \, | \, x, \, g)$ that measures how likely $y$ is to be the $g$-transformed version of $x$, we can consider the posterior over transformations $p(g \, | \, x, \, y)$.
Typically, the posterior turns into a complicated and intractable distribution, but we show that if the prior is a harmonic density and the likelihood is Gaussian, the posterior distribution is again a harmonic density whose parameters are easily obtained using the generalized FFT algorithm.
Furthermore, the global mode of this posterior (the optimal transformation) can be computed efficiently by performing yet another FFT.

In this paper we provide both an abstract treatment of the theory of harmonic densities that covers a fairly broad class of manifolds in a uniform and easily understandable manner, and a concrete instantiation of this theory in the case of the rotation groups $\SO2$ and $\SO3$, and their homogeneous spaces, the circle $S^1$ and sphere $S^2$.

The rest of this paper is organized as follows.
We begin by discussing related work in section \ref{sec:related_work}, followed by a brief summary of non-commutative harmonic analysis
for a machine learning audience in section \ref{sec:preliminaries}.
In section \ref{sec:harmonic_exponential_families}, we define harmonic exponential families and present an FFT-based maximum-likelihood estimation algorithm.
In section \ref{sec:harmonic_densities_as_conjugate_priors}, we show that harmonic exponential families are the conjugate priors in the Bayesian transformation estimation problem, and present an efficient MAP inference algorithm.
Our earthquake modelling experiments are presented in section \ref{sec:experiments}, followed by a discussion and conclusion.

\section{Related Work}
\label{sec:related_work}

The harmonic exponential families were first defined abstractly (but not named) by \citet{Diaconis1988}.
Despite their long history and a significant body of literature devoted to them (see \citet{Mardia1999}), this class of densities has remained intractable until now for all but the simplest cases.

In fact, various commonly used distributions on the circle and the sphere are harmonic densities of low degree.
Among these are the 2-parameter von-Mises distribution on the circle and the 5-parameter Kent distribution on the sphere.
These are both exponential families, about which \citet{Mardia1999} write (p. 175):
``Although exponential models have many pleasant inferential properties, the need to evaluate the normalizing constant (or at least the first derivative of its logarithm) can be a practical difficulty.''

What is a practical difficulty for the unimodal distributions with few parameters mentioned above becomes a showstopper for more flexible exponential families with many parameters.
The harmonic exponential family for the circle is known as the generalized von-Mises distribution, and can be defined for any band-limit / order / degree $L$.
However, no scalable maximum-likelihood estimation algorithm is known.
\citet{Gatto2007} (who work only with the 4-parameter $L=2$ distribution) compute the normalizing constant by a truncated infinite sum, where each term involves an expensive Bessel function evaluation.

\citet{Beran1979} studies an exponential family that is equivalent to the harmonic exponential family on the sphere, but does not provide a scalable learning algorithm.
At each iteration, the proposed algorithm computes all $O(L^2)$ moments of the distribution by numerical integration.
This requires $O(L^2)$ samples per integration, making the per-iteration cost $O(L^4)$.
Beran further suggests using a second order optimization method, which would further increase the per-iteration cost to $O(L^6)$.

This is clearly not feasible when $L$ is measured in the hundreds, and parameter counts in the 10's of thousands, as is needed in the experiments reported in section \ref{sec:experiments}.
The algorithm described in this paper is simple, generic across manifolds, and fast (per-iteration complexity $O(L^2 \log^2 L)$ in the spherical case, for instance).
It can be applied to any manifold %compact homogeneous space or Lie group
for which an FFT has been developed.

\section{Preliminaries}
\label{sec:preliminaries}

The manifolds we consider in this paper are either Lie groups or closely related manifolds called homogeneous spaces, and the sufficient statistics that we use come from harmonic analysis on these manifolds.
Since these concepts are not widely known in machine learning, we will review them in this section.
For more details, we refer the reader to \citet{Chirikjian2001, Sugiura1976, Kondor2008, Goodman2009}.

\subsection{Lie Groups}

A \emph{transformation group} $G$ is a set of invertible transformations that is closed under composition and taking inverses: for any $g, h \in G$, the composition $g h$ is again a member of $G$, and so are the inverses $g^{-1}$ and $h^{-1}$.

A \emph{Lie group} is a group that is also a differentiable manifold.
For example, the group $\SO3$ of 3D rotations is a Lie group.
It can be represented as a set of matrices,
\begin{equation}
  \SO{3} = \{ R \in \R^{3\times 3} \, | \, RR^T = I, \, \det(R) = 1\},
\end{equation}
so one way to think about \SO3 is as a 3-dimensional manifold embedded in $\R^{3\times 3}$.
For technical reasons, we will further restrict our attention to \emph{compact Lie groups}, i.e. Lie groups that are closed and bounded.

\subsection{Harmonic analysis on compact Lie groups}
\label{sec:harmonic_analysis_on_compact_lie_groups}

The basic idea of the generalized Fourier transform on compact Lie groups is to expand a function $f : G \rightarrow \C$ as a linear combination of carefully chosen basis functions.
These basis functions have very special properties, because they are the matrix elements of \emph{irreducible unitary representations} (IURs) of $G$.
These terms will now be explained.

A \emph{representation} of a group $G$ on a vector space $V$ is a map $R$ from the group to the set of invertible linear transformations of $V$ that preserves the group structure in the following sense:
\begin{equation}
  \label{eq:representation}
  R(gh) = R(g) R(h).
\end{equation}
Note that $gh$ denotes composition of group elements while $R(g) R(h)$ denotes matrix multiplication (once we choose a basis for $V$, that is).

In computer vision, we encounter group representations in the following way.
An image is represented as a vector $x \in \mathbb{R}^n$ of pixel intensities, and to be concrete, we take $G$ to be the group $\SO2$ consisting of rotations of the plane.
Then $R(\theta)$ is the matrix such that $R(\theta)x$ is the image $x$ rotated by angle $\theta$.

As this example shows, many representations do not change the norm of the vectors on which they act: $\forall g \in G, \forall x \in V: \|U(g) \, x\| = \|x\|$.
Such representations are called \emph{unitary}.
Unitary representations tend to be easier to work with both analytically and numerically.

Given a unitary representation $U$ and a unitary matrix $F$, one can define an equivalent representation $T(g) = F^{-1} U(g) F$.
In computer vision one can think of $F$ as a matrix containing image features in the rows.
One can now try to find $F$
such that for every $g$, the matrix $F^{-1} U(g) F$ is block diagonal with the same block structure.
If we continue to block-diagonalize until no further diagonalization is possible, we end up with blocks called \emph{irreducible unitary representations}\footnote{Technically, we have defined the slightly easier to understand notion of \emph{indecomposability}, which in this context implies irreducibility.}.
The IURs of a compact group can be indexed by a discrete index $\lambda$, and we denote the IURs as $U^\lambda(g)$.

As an example, consider the 2D rotation group \SO2.
Since $\SO2$ is commutative, its representation matrices can be jointly diagonalized and so
the IURs of $\SO2$ are $1 \times 1$ matrices:
\begin{equation}
  U^\lambda_{00}(g) = e^{i \lambda g}.
\end{equation}
They satisfy the composition rule $e^{i\lambda(g + h)} = e^{i\lambda g} e^{i\lambda h}$, which is the manifestation of eq. \ref{eq:representation} for this representation.
The standard Fourier series of a function on the circle is an expansion in terms of these matrix elements, which shows that standard Fourier analysis is a special case of the more general transform to be defined shortly.

The matrix elements of IURs of $\SO3$ are known as Wigner D-functions.
They are defined for $\lambda = 0, 1, 2, \ldots$ and $-\lambda \leq m, n \leq \lambda$, so the irreducible representations are $(2 \lambda + 1)$-dimensional.
Wigner D-functions can be expressed as sums over products of sinusoids or complex exponentials \cite{Pinchon2007},
but the formulae are somewhat unwieldy so that it is easier to think only about their general properties.

The most important general property of the matrix elements of IURs is that they are orthogonal:
\begin{equation}
  \begin{aligned}
    \langle U^\lambda_{mn}(g), \, U^{\lambda'}_{m'n'}(g) \rangle
    &\equiv
    \int_G U^\lambda_{mn}(g) \, \overline{U^{\lambda'}_{m'n'}(g)} \, d\mu(g) \\
    &=
    \frac{\delta_{\lambda \lambda'} \delta_{mm'} \delta_{nn'}}{\dim \lambda}.
  \end{aligned}
\end{equation}
Here $\mu$ is the normalized Haar measure, which is the natural way to measure volumes in $G$ \cite{Sugiura1976}, and $\dim \lambda$ is the dimension of the representation.
One can verify that the complex exponentials $e^{i \lambda g}$ are indeed orthonormal with $d\mu(g) = \frac{dg}{2 \pi}$ and $\dim \lambda = 1$.

Intuitively, the matrix elements are like a ``complete orthogonal basis'' for the space $L^2(G)$ of square integrable functions on $G$.
That is, it can be proven that any function $f \in L^2(G)$ can be written as
\begin{equation}
  \label{eq:inverse_fourier}
  f(g) = \sum_\lambda \sum_{mn} \eta^\lambda_{mn} T^\lambda_{mn}(g) \equiv \left[ \mathcal{F}^{-1} \eta\right](g),
\end{equation}
where $T^\lambda_{mn}(g) = \sqrt{\dim \lambda} \, U^\lambda_{mn}(g)$ are the $L^2$-normalized matrix elements and $\eta$ are the Fourier coefficients of $f$.

Integrating eq. \ref{eq:inverse_fourier} against a matrix element and using orthonormality, we find:
\begin{equation}
  \label{eq:fourier_coefficients}
  \eta^\lambda_{mn} = \int_G f(g) \overline{T^\lambda_{mn}(g)} d\mu(g).
\end{equation}
This is the (generalized) Fourier transform for compact groups, denoted
$$\eta = \mathcal{F} f.$$

Fast and exact algorithms for the computation of Fourier coefficients from samples of bandlimited functions on the rotation groups \SO2 and \SO3 have been developed, and the theory required to construct such algorithms for general compact Lie groups is understood \cite{Maslen1997}.
The group \SO2 is isomorphic to the circle, so for $G = \SO2$ equation \ref{eq:inverse_fourier} reduces to a standard Fourier series on the circle, for which the well-known $O(L \log L)$ FFT algorithm can be used.
The \SO3 FFT has complexity $O(L^3 \log^2 L)$ for bandlimit $L$
\cite{Maslen1997, Kostelec2008, Potts2009}.
This is a tremendous speedup compared to the naive $O(L^6)$ algorithm,
and the algorithms presented in this paper would certainly not be feasible without the FFT.

In section \ref{sec:harmonic_exponential_families} we discuss how these generalized FFT algorithms can be used to efficiently compute moments, but first we discuss the generalization of the Fourier transform on compact Lie groups to the Fourier transform on certain manifolds that are not groups.

\subsection{Harmonic analysis on homogeneous spaces}
\label{sec:harmonic_analysis_on_homogeneous_spaces}

A \emph{homogeneous space} for a Lie group $G$ is a manifold $H$ such that for any two points $x, \, y \in H$ we can find a transformation $g \in G$ with $g x = y$.
For example, the plane is a homogeneous space for the translation group, and the sphere is a homogeneous space for the 3D rotation group.
The plane is not a homogeneous space for the 2D rotation group, because points at different radii cannot be rotated into each other.

If we pick an origin $o \in H$, such as the north pole of the sphere, we can identify any other point $h \in H$ by specifying how to transform the origin to get there: $h = g o$.
This identification will not be unique, though, if there is a nontrivial subgroup $K$ of $G$ containing transformations that leave the origin invariant: $K = \{ k \in G \, | \, k o = o \}$.
This is because if $h = g o$, then also $h = gk o$, so both $g$ and $gk$ identify $h$.
On the sphere for example, we can transform the north-pole into a point $h$ by first doing an arbitrary rotation around the north-pole axis (which leaves the north-pole unchanged) and then rotating the result to $h$.

Hence, one can think of the \emph{points} $h = go$ in a homogeneous space $H$ as \emph{sets} $g K = \{g k \, | \, k \in K\}$ (called cosets) of group elements that are equivalent with respect to their effect on an arbitrarily chosen origin $o$ of $H$.
It follows that one can think of functions on a homogeneous space as functions on the group, with the special property that they are (right) invariant to transformations from $K$: %$f(gk) = f(g)$),
\begin{equation}
  f(gk) = f(g) \;\; \forall g \in G, k \in K,
\end{equation}
because right-multiplication by $k$ will only shuffle the elements within each coset.
Finally, one can show
(see supplementary material)
that in a suitable basis, a subset of the matrix elements of IURs form a basis for the linear space of square-integrable functions on $G$ with this invariance property, which allows us to define the Fourier transform also for functions on $H$.

The exact same equations (\ref{eq:fourier_coefficients} and \ref{eq:inverse_fourier}) that define the Fourier and inverse Fourier transform for a compact Lie group, define these transforms for a compact homogeneous space,
but only a subset of the coefficients $\eta^\lambda_{mn}$ will be non-zero.
For the sphere, the matrix elements $T^\lambda_{m0}$ are equal\footnote{Various normalization and phase conventions are in use for the spherical harmonics, but it is enough to know that $Y^{\lambda}_m \propto T^{\lambda}_{m0}$.}
to the spherical harmonics $Y^\lambda_m$, which form a basis for $L^2(S^2)$.
Fast spherical Fourier transform algorithms were developed by \citet{Driscoll1994}.

\subsection{Exponential families}
An exponential family is a class of densities of the form:
\begin{equation}
  \label{eq:expfam}
  p(g \, | \, \eta) = \frac{1}{Z_\eta} \exp{\left( \eta \cdot T(g) \right)}.
\end{equation}
It is determined by a choice of \emph{sufficient statistics} $T$, that take the random variable $g$ and produce a vector of real statistics $T(g)$.

To learn the parameters $\eta$, one can perform gradient-based optimization of the log-likelihood of a set of iid samples $g_1, \ldots, g_N$.
The gradient is the moment discrepancy:
\begin{equation}
  \label{eq:logp_grad}
  \nabla_\eta \left( \frac{1}{N} \sum_{i=1}^N \ln p(g_i \, | \, \eta)\right)
  =
  \bar{T} - \mathbb{E}_{p(g|\eta)}[T(g)],
\end{equation}
where $\bar{T} = \frac{1}{N} \sum_{i=1}^N T(g_i)$ are the empirical moments. 
There is generally no closed form for the analytical moments (the expectation in eq. \ref{eq:logp_grad}), so a numerical approximation is needed.

\section{Harmonic Exponential Families}
\label{sec:harmonic_exponential_families}

We define a harmonic exponential family on a group or homogeneous space as an exponential family where the sufficient statistics are given by a finite number of matrix elements of IURs.
This makes sense only if the function $\eta \cdot T(g)$ is real-valued, so that it can be interpreted as an unnormalized log-probability.
The easiest way to guarantee this is to take $\eta$ to be real, and to use real functions obtained as a sparse linear combination of complex matrix elements $T(g)$ as sufficient statistics.
For example, $\frac{1}{2} (e^{i \lambda \theta} + e^{-i\lambda \theta}) = \cos(\lambda \theta)$.
From here on, we take $T$ to be real, $L^2$-normalized functions and $\mathcal{F}$ the expansion of a real function in terms of real basis functions $T$.

The key observation that leads to an efficient algorithm for computing the moments of a harmonic density is that
the moments of such a density are its Fourier coefficients:
\begin{equation}
  \begin{aligned}
    \mathbb{E}_{p(g|\eta)}\left[ T(g) \right]
    =
    \int_G p(g | \eta) \, T(g) \, d\mu(g)
    =
    \mathcal{F} \, p.
  \end{aligned}
\end{equation}
Hence, one can obtain all $J$ moments at once by sampling $p$ on a finite grid and then computing its Fourier transform using a fast algorithm.
As explained in section \ref{sec:approximation_quality}, the discretization error can be made extremely small using only $O(J)$ spatial samples.

However, even evaluating $p$ at a single position takes $O(J)$ computations when using $J$ sufficient statistics so that the overall complexity is still $O(J^2)$.
Furthermore, in order to evaluate $p$ we need to know the normalizing constant $Z_\eta$.

The following derivation shows that we can work with the unnormalized density $\varphi(g \, | \, \eta) = \exp{(\eta \cdot T(g))}$ instead:
\begin{equation}
  \begin{aligned}
    \lbrack \mathcal{F} p \rbrack^\lambda_{mn}
    &=
    \int_G p(g \, | \, \eta) \, T^\lambda_{mn}(g) \, d\mu(g) \\
    &=
    \frac{1}{Z_\eta} \int_G \varphi(g \, | \, \eta) \, T^\lambda_{mn}(g) \, d\mu(g) \\
    &=
    \frac{[\mathcal{F} \varphi]^\lambda_{mn}}
         {[\mathcal{F} \varphi]^0_{00}},
  \end{aligned}
\end{equation}
The last step uses the fact that $T^0_{00}(g) = 1$ so that $[\mathcal{F} \varphi]^0_{00}$ is equal to the normalizing constant:
\begin{equation}
  \begin{aligned}
    \left[\mathcal{F} \varphi \right]^0_{00}
    &=
    \int_G \varphi(g \, | \, \eta) \, T^0_{00}(g) \, d\mu(g)
    =
    Z_\eta
  \end{aligned}
\end{equation}

Next, observe that we can evaluate $\ln \varphi$ efficiently at $O(J)$ spatial points using the inverse FFT:
\begin{equation}
  \ln \varphi(g \, | \, \eta) = \eta \cdot T(g) = \left[ \mathcal{F}^{-1} \eta \right](g)
\end{equation}
This computation is exact, because the log-density is bandlimited (i.e. there are only finitely many parameters).
Element-wise exponentiation then gives us $\varphi$ evaluated on a grid.

So we have an efficient algorithm for computing moments:
\begin{enumerate}
\item Compute $\varphi = \exp{(\mathcal{F}^{-1} \eta)}$.
  \item Compute $M = \mathcal{F} \, \varphi$
  \item Compute $\mathbb{E}_{p(g|\eta)}\left[T(g)\right] = M / M^0_{00}$.
\end{enumerate}

To make this computation numerically stable for highly peaked densities, one should apply the ``log-Fourier-exp'' trick described in the supplementary material.

\subsection{Approximation quality}
\label{sec:approximation_quality}

Even though the Fourier coefficients are defined as definite integrals (eq. \ref{eq:fourier_coefficients}), the discrete FFT algorithms compute \emph{exact} Fourier coefficients, provided the function from which the discrete samples were gathered is \emph{bandlimited}.
A function is bandlimited if the coefficients $\eta^\lambda_{mn}$ are zero for $\lambda$ greater than the band-limit $L$.
Although the function $\ln \varphi(g) = \eta \cdot T(g)$ is bandlimited, the function $\varphi(g) = \exp{(\eta \cdot T(g))}$ is not, so the computed coefficients are not exactly equal to the Fourier coefficients of $\varphi$.

However, the function $\varphi(g)$ is smooth (infinitely differentiable), and a standard result in Fourier analysis shows that the spectrum of a smooth function decays to zero asymptotically faster than $O(1/\lambda^n)$ for any $n$.
So our function will be ``effectively bandlimited'', in the sense that coefficients for $\lambda$ greater than some pseudo-bandlimit will have negligible values.
If $L$ is the maximum degree of the sufficient statistics (the bandlimit of $\eta \cdot T(g)$),
one can obtain near-exact moments by computing the FFT up to the pseudo-bandlimit $\alpha L$ for some oversampling factor $\alpha$.
In practice, we use values for $\alpha$ ranging from $2$ to $5$.

\section{Harmonic Densities as Conjugate Priors}
\label{sec:harmonic_densities_as_conjugate_priors}

In this section we discuss the Bayesian transformation inference problem, where the goal is to infer a posterior over a Lie group of transformations given only a set of correspondence pairs (such as images before and after a camera motion).
It turns out that the harmonic exponential families are the conjugate priors for this problem, and again, the generalized FFT is key to performing efficient inference.

The observed data in the Bayesian transformation inference problem are pairs of vectors $(x, \, y)$ in $\R^D$ that could represent images, space-time blocks of video, point-clouds, optical-flow fields, fitted geometric primitives, parameters of a function or other objects.
In order to infer anything about a latent transformation $g$, we must know the group representation $R(g)$ that acts on the observed data.
If our data is an image $x : \mathbb{R}^2 \rightarrow \mathbb{R}$, we get a representation on the Hilbert space in which $x$ lives: $[R(g) x](p) = x(g^{-1} p)$, where $p$ is a point in the plane.
In the finite-dimensional analogue, where $x$ is a vector of pixel intensities, $R(g)$ will be close to a permutation matrix that takes each pixel to its proper new position.
For compact groups\footnote{This more generally true for unimodular groups.} this representation is unitary, and this is what we will assume for $R(g)$ from now on.
If the representation is not known in advance, it can also be learned from data \cite{Cohen2014}.

If we assume that observation $x$ is the $g$-transformed version of $y$ with some independent Gaussian noise with variance $\sigma^2$ added, the likelihood function is given by
\begin{equation}
  \label{eq:transformation_likelihood}
  p(x \, | \, y, g) = \mathcal{N}(x \, | \, R(g) y, \sigma^2).
\end{equation}

As discussed in section \ref{sec:harmonic_analysis_on_compact_lie_groups}, we can bring $R$ in block-diagonal form by a unitary change of basis: $U(g) = F^{-1} R(g) F$.
The matrix $U$ is block diagonal, and the blocks
$U^\lambda$ are equal to the $L^2$-normalized sufficient statistics $T^\lambda$ up to a scale factor:
$T^\lambda(g) = \sqrt{\dim \lambda} \, U^\lambda(g)$.
To simplify the computations, we shall work with data in this new basis: $\hat{x} = F x$ so that $p(\hat{x} \, | \, \hat{y}, \, g) = \mathcal{N}(\hat{x} \, | \, U(g) \hat{y}, \sigma^2)$.

If we now choose as prior $p(g)$ a member of the harmonic exponential family on $G$, the posterior $p(g \, | \, \hat{x}, \, \hat{y})$ is of the same form as the prior:
\begin{equation}
  \label{eq:posterior}
  \begin{aligned}
    p(g \, | \, \hat{x}, \, \hat{y})
    &\propto
    p(\hat{x} \, | \, \hat{y}, \, g) p(g) \\
    &\propto
    \exp{\left(-\frac{1}{2\sigma^2} \|\hat{x} - U(g) \hat{y}\|^2 + \eta \cdot T(g) \right)} \\
    &\propto
    \exp{\left(\frac{1}{\sigma^2} \hat{x}^T U(g) \hat{y} + \eta \cdot T(g) \right)} \\
    &=
    \exp{\left( \sum_\lambda \left( \eta^\lambda + \frac{\hat{x}_\lambda \, \hat{y}_\lambda^T}{\sigma^2 \sqrt{\dim \lambda}} \right) \cdot T^\lambda(g) \right)} \\
  \end{aligned}
\end{equation}
i.e. we have a conjugate prior.
The derivation relies on the unitarity of the representation: in expanding $\|\hat{x} - U(g) \hat{y}\|^2$, we find a term $\|U(g) \hat{y} \|^2$ which is equal to $\|\hat{y} \|^2$, making the dependence on $U(g)$ \emph{linear} (as it is in the prior).

\subsection{Example: Bayesian analysis of camera rotation}

To make matters concrete, we show how to compute a posterior over the rotation group $\SO3$ given two images taken before and after a camera rotation.
An image is modeled as a function $x : S^2 \rightarrow \R$
on the sphere, so that a camera rotation $g \in \SO3$ acts by rotating this function over the sphere: $[R(g) x](p) = x(g^{-1}p)$.
We parameterize points $p \in S^2$ as $p = (\varphi, \theta)$ for $\varphi \in [0, 2 \pi]$ and $\theta \in [0, \pi]$.
Recall from section \ref{sec:harmonic_analysis_on_homogeneous_spaces} that we can represent $p \in S^2$ by a coset representative $g_p \in \SO3$, which we parameterize using Euler angles as $g_p = (\varphi, \theta, 0)$.
The transformation $g_p$ takes the origin of the sphere to $p$.

It is well known\footnote{The derivation can be found in the supplementary material.} that in this context the matrix $F$ --- defined in the previous section as the matrix that block-diagonalizes the representation $R$ --- is given by the spherical Fourier transform $\mathcal{F}$, which can be computed by an FFT.
This means that if we represent $x$ by its Fourier coefficients $\hat{x} = \mathcal{F} x$, the coefficients of the rotated function $R(g)x(p) = x(g^{-1} p)$ are given by $U(g) \hat{x}$, where $U(g)$ is a block-diagonal matrix with irreducible representations $U^\lambda$ as blocks.
As derived in the previous section, the parameters of the posterior can easily be obtained in this basis by a block-wise outer product.

\begin{figure}
  \centering
  \subfloat{
    \includegraphics[width=0.23\textwidth]{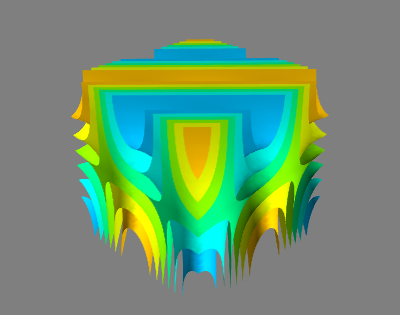}
  }
  \subfloat{
    \includegraphics[width=0.23\textwidth]{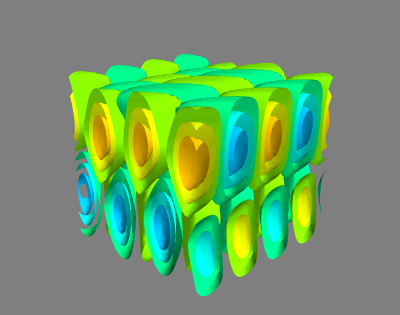}
  } \\
  \subfloat{
    \includegraphics[width=0.1\textwidth]{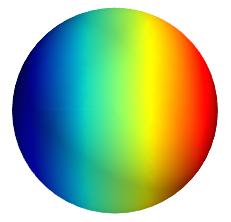}
    \includegraphics[width=0.1\textwidth]{SH1_1}
  }
  \subfloat{
    \includegraphics[width=0.1\textwidth]{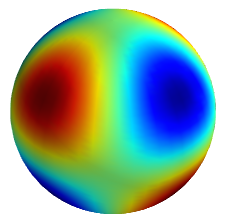}
    \includegraphics[width=0.1\textwidth]{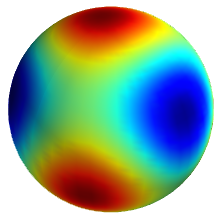}
  }
  \caption{Posterior distributions (top) for two correspondence pairs (bottom).}
  \label{fig:posteriors}
\end{figure}

Figure \ref{fig:posteriors} shows the posterior $p(g ,\ | \, x, \, y)$ for two synthetic spherical images $x^1$ and $x^2$, and their rotations $y^1 = x^1$ (no rotation) and $y^2 = U(0, \pi / 3, \pi / 2) x^2$.
The posterior is plotted as a 3D isocontour in the ZYZ-Euler angle parameter space $(\alpha, \beta, \gamma) \in [0, 2 \pi] \times [0, \pi] \times [0, 2\pi]$.
Although this plot looks like a box, the actual manifold has the topology of a projective 3-sphere.
By construction, the density is spread through the parameter space in a way that is consistent with this topology: no discontinuities arise where wraparound in the parameter space occurs.
This is desirable, because the wraparound is not an intrinsic property of the manifold.

Due to the symmetry of these figures certain rotations cannot be distinguished from the ``true'' rotation, so that the modes of the posterior distributions are supported on an entire subgroup of $\SO3$.
Real images will not have such a high degree of symmetry, but it will nevertheless often be the case that a unique optimal transformation does not exist \cite{Ma1999Euclidean}.
Indeed, current keypoint based transformation estimation methods can easily get confused by repeating structures in an image, such as several identical windows on a building.
Although more experimental work is needed, our method has the theoretical advantage that besides keypoints (which form a group representation), it can make use of parts of the image that do not allow for keypoints to be reliably placed (such as edges), while always providing a truthful impression of the degree to which a unique transformation or subgroup can be identified form the data.

\subsection{MAP inference}
As Bayesians we are done here, but for some applications one may wish to find a single most likely transformation.
To find the optimal transformation, first perform posterior inference (steps 1 and 2) and then maximize (step 3):
\begin{enumerate}
  \item Compute $\hat{x} = \mathcal{F} \,x$ and $\hat{y} = \mathcal{F} \, y$.
  \item Compute $\bar{\eta}^\lambda = \eta^\lambda + \frac{1}{\sigma^2 \sqrt{\dim \lambda}} \hat{x}_\lambda \hat{y}_\lambda^T$
  \item Compute $g^* = \arg\max_i [\mathcal{F}^{-1} \bar{\eta}](g_i)$
\end{enumerate}
The $\arg\max$ ranges over all the points in a finite grid on $G$ used by the FFT synthesis $\mathcal{F}^{-1}$.
Optionally, one can refine the optimum $g^*$ by performing a few steps of gradient-based optimization on $\bar{\eta} \cdot T(g)$ to get sub-pixel accuracy.

\section{Experiments}
\label{sec:experiments}

\subsection{Modelling the spatial distribution of earthquakes}

We compare our model and MLE algorithm to a Kent Mixture Model (KMM) on the problem of modelling the spatial distribution of significant earthquakes on the surface of the earth.

We obtained the Significant Earthquake Dataset \cite{NGDC} from the National Geophysical Datacenter of the National Oceanographic and Athmospheric Administration.
In total, the dataset contains $5780$ earthquakes with complete information on the position of their epicenter, and $53$ earthquakes whose coordinates are missing (these were discarded in our experiments).
We did not model the severity of the earthquake, but only the occurrence of significant earthquakes (as defined by \cite{NGDC}).

\subsubsection{Mixture of Kent distributions}

The 5-parameter Kent distribution \cite{Mardia1999} is the spherical analogue of the normal distribution with unconstrained covariance.
Being unimodal, the Kent distribution is not flexible enough to describe complicated distributions such as the spatial distribution of earthquakes.
The most flexible distribution on the sphere that we have found in the literature is the Kent Mixture Model, first described by \citet{Peel2001}.
The KMM is trained using the EM algorithm.
We use the open source Python implementation of the EM algorithm for KMMs by \citet{Hofer2014}.

Unlike the harmonic densities, the log-likelihood of this model is not convex and contains many singularities where a mixture component concentrates on a single data point and decreases its variance indefinitely.
For this reason, we perform randomly initialized restarts until the algorithm has found 10 non-degenerate solutions, of which we retain the one with the best cross-validation log-likelihood.
No regularization was used, because for the models that could be trained within a reasonable amount of time, no overfitting was observed.

\subsubsection{The $S^2$ harmonic density}

The harmonic density on the 2-sphere uses spherical harmonics as sufficient statistics.
The empirical moments are easily computed using standard spherical harmonic routines,
but we found that for high orders the SciPy routines are slow and numerically unstable.
The supplementary material describes a simple, fast, and stable method for the evaluation of spherical harmonics.
The computation of spherical harmonics up to band-limit $L = 200$ (for a total of $(L+1)^2 = 40401$ spherical harmonics) for $5780$ points on the sphere took half a minute using this method and is performed only once for a given dataset.

For regularization we use a diagonal Gaussian prior on $\eta$, where the precision $\beta^\lambda_m$ corresponding to the coefficient of $Y^\lambda_m$ is given by $\beta^\lambda_m = \alpha \dim \lambda = \alpha (2 \lambda + 1)$ (for some regularization parameter $\alpha$).
This scheme is inspired by the fact that $\dim \lambda$ is the discrete Plancherel measure \cite{Sugiura1976}, and the empirical observation that the fitted coefficients become approximately uniform when weighted as $\eta^\lambda_m \, \sqrt{\dim \lambda}$.
Note that adding regularization does not change the convexity of the objective function.

To find maximum a posteriori parameters $\hat{\eta}$ for the spherical harmonic density, we perform iterative gradient-based optimization on the log-posterior.
The gradients (moment discrepancies) are computed using the FFT-based method described in section \ref{sec:harmonic_exponential_families}.
We use the spherical FFT algorithm implemented in the NFFT library \cite{Keiner2009, Kunis2003}.
The gradients are fed to a standard implementation of the L-BFGS algorithm.

\subsubsection{Results}

\begin{figure}[t]
  \centering
  \includegraphics[width=0.5\textwidth]{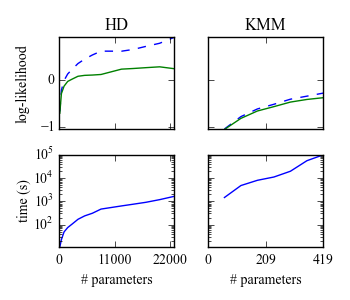}
  \caption{Top: cross-validation train (dashed line) and test (solid line) log-likelihood for the Harmonic Density (HD) and Kent Mixture Model (KMM). Bottom: number of parameters versus training time for both models.}
  \label{fig:cv}
\end{figure}

Figure \ref{fig:cv} shows the average train and test log-likelihood over 5 cross-validation folds, for the spherical harmonic density and the mixture of Kent distribution.
The plotted values correspond to the regularization settings that yielded the best test log-likelihood.

\begin{figure}
  \centering
  \subfloat[Significant Earthquake Dataset]{
    \includegraphics[width=0.25\textwidth]{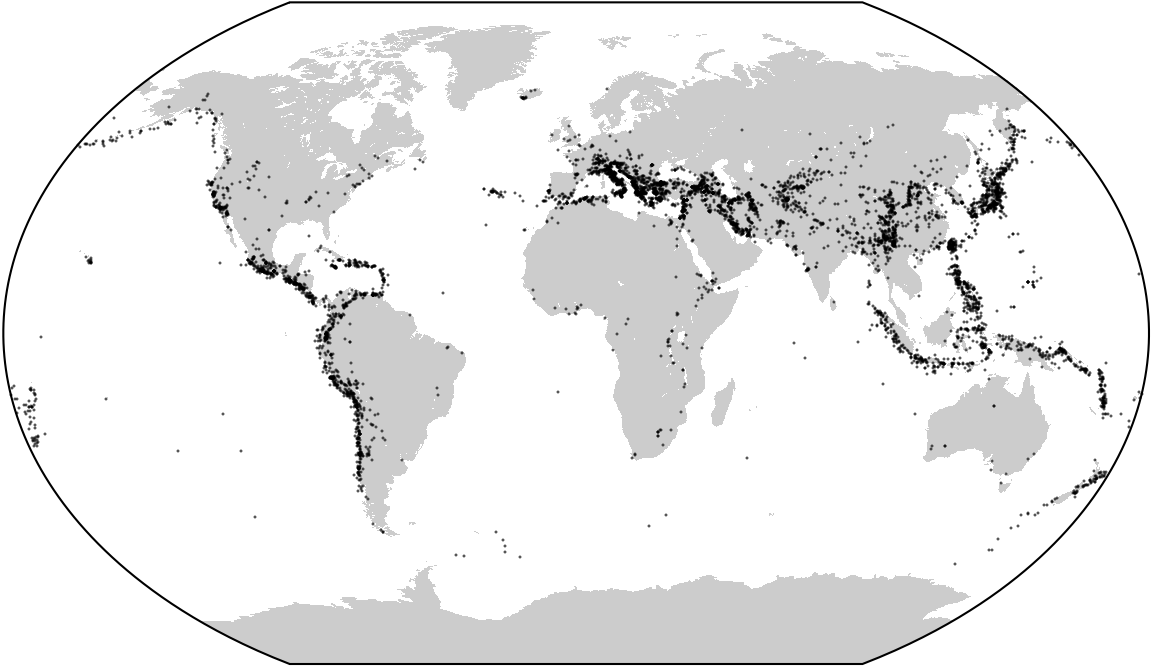}
  } \\
  \subfloat[Harmonic Density]{
    \includegraphics[width=0.23\textwidth]{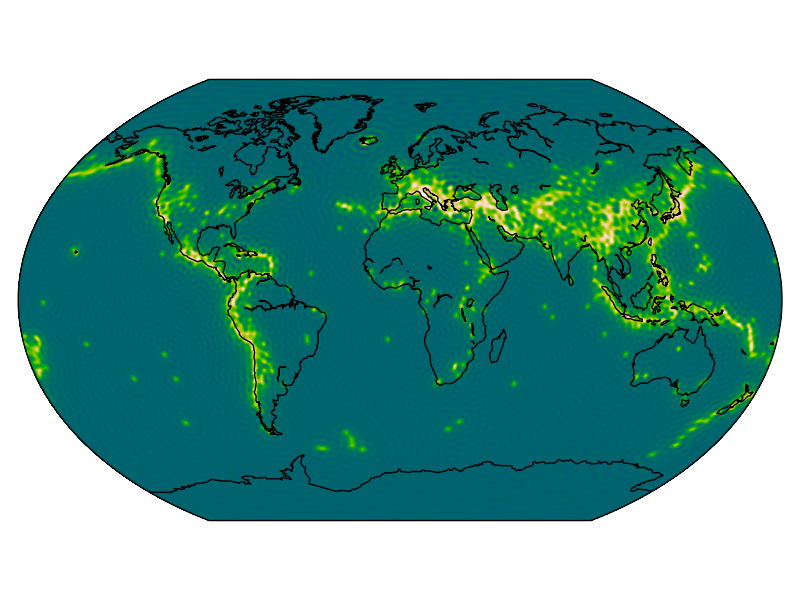}
  }
  \subfloat[Kent Mixture Model]{
    \includegraphics[width=0.23\textwidth]{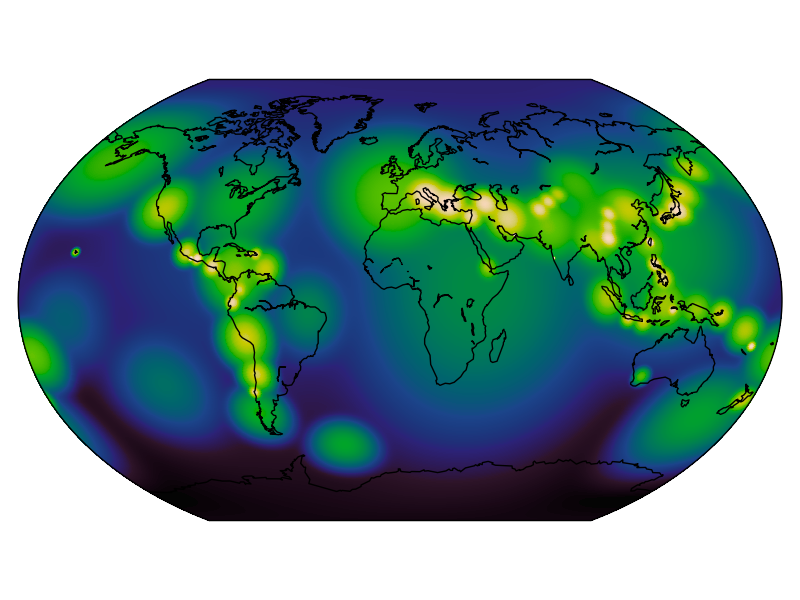}
  }
  \caption{Log-probability for harmonic density and Kent mixture model, plotted using a perceptually accurate linear-lightness colormap on the same intensity scale.}
   \label{fig:maps}
\end{figure}

The KMM reached an average test log-likelihood of $-0.37$ (with standard deviation of $0.03$ over 5 cross-validation folds) using $70$ mixture components ($5 \times 70 + 69 = 419$ parameters).
The harmonic density reached an average test log-likelihood of $+0.3$ (with standard deviation of $0.036$ over 5 cross-validation folds), using bandlimit $L = 140$ ($19880$ parameters).
The HD still outperforms the KMM when given a similar number of parameters: for $L = 20$ ($440 \approx 419$ parameters), the log-likelihood is $-0.28 > -0.37$, with standard deviation $0.028$, and for $L = 19$ ($399$ parameters), the log-likelihood is $-0.3$ with standard deviation $0.03$.
The dataset and the learned densities are plotted in figure \ref{fig:maps}, clearly showing the superiority of the harmonic density.

\section{Discussion and Future Work}

What could explain the difference in log-likelihood between our model and the KMM?
We believe two interrelated factors are driving this difference: the expressiveness of the model and the ease of optimization.
Leaving technicalities aside, it is clear that both the spherical harmonic exponential family and the KMM can approximate any well-behaved density, given enough parameters and an optimization oracle.
However, as is clear from figure \ref{fig:cv}, training time becomes prohibitive for the KMM for more than $70$ mixture components / $419$ parameters, while the harmonic density can efficiently be fit for tens of thousands of parameters.

Furthermore, the KMM training algorithm (EM) can easily get stuck in local optima, or converge on a degenerate solution.
This is the main reason for the poor runtime performance; while the KMM code could be further optimized,
it is the fact that so many restarts are required to find a good fit that makes the algorithm slow.
The log-likelihood function of the harmonic density, on the other hand, is convex, and the L-BFGS optimizer will typically converge to the global optimum in some $20-100$ iterations.

As can be seen in figure \ref{fig:maps}, the harmonic density produces slight ringing artifacts that can be seen only in a log-plot such as this.
These are the result of the limited bandwidth of the log-density, and will become progressively less pronounced as the number of parameters is increased.
While they are clearly visible in log-space, the actual difference between peaks and valleys is on the order of $10^{-3}$ for bandlimit $L = 100$.
The artifacts are not visible on a non-logarithmic plot (and in such a plot the KMM density is hardly visible at all when plotted on the same intensity scale as the harmonic density, because the peaks are much lower).
The harmonic density also tends to prefer heavier tails, which is probably accurate for many problems.

An interesting direction for future work is the extension to non-compact groups.
While the mathematical theory becomes much more technical for such groups, 
\cite{Kyatkin2000} have already succeeded in developing FFT algorithms for the Euclidean motion group which is non-compact.
From there it should be relatively straightforward to develop harmonic densities on the Euclidean group.
Harmonic densities on the Euclidean, affine or even projective group should find many applications in robotics and computer vision (see e.g. \cite{Kyatkin1999a}).

\section{Conclusion}

We have studied a class of exponential families on compact Lie groups and homogeneous manifolds, which we call harmonic exponential families.
We have shown that for these families, maximum likelihood inference, posterior inference and mode seeking can be implemented using very efficient generalized Fast Fourier Transform algorithms.
In the Bayesian setting, we have shown that harmonic exponential families appear naturally as conjugate priors in the generic transformation inference problem.
Our experiments show that training harmonic densities is fast even for very large numbers of parameters, and that far superior likelihood can be achieved using these models.

\section*{Acknowledgements}

This research was supported by NWO (grant number NAI.14.108), Facebook and Google.

\bibliography{ICML2015}
\bibliographystyle{icml2015}

\end{document}